\documentclass{article}

\usepackage[final,nonatbib]{neurips_2023}

\usepackage[utf8]{inputenc} 
\usepackage[T1]{fontenc}    
\usepackage{hyperref}       
\usepackage{url}            
\usepackage{booktabs}       
\usepackage{amsfonts}       
\usepackage{nicefrac}       
\usepackage{microtype}      
\usepackage{xcolor}         
\usepackage{tcolorbox}
\usepackage{amsmath,amssymb,amsfonts}
\usepackage[scientific-notation=true]{siunitx}
\usepackage{adjustbox}
\usepackage{array}
\newcolumntype{P}[1]{>{\centering\arraybackslash}p{#1}}
\usepackage{chemfig}
\usepackage{soul}  
\usepackage{romannum}
\usepackage{ltablex}
\usepackage{xcolor}
\newcolumntype{b}{X}
\newcolumntype{s}{>{\hsize=2.2\hsize}X}

\title{Crossing New Frontiers: Knowledge-Augmented Large Language Model Prompting for Zero-Shot Text-Based De Novo Molecule Design}
\vspace{-4mm}
\author{
  Sakhinana Sagar Srinivas\thanks{Designed and programmed research, conducted experiments, analyzed results, and drafted manuscript}, \textbf{Venkataramana Runkana}\\
  TCS Research \\
  \texttt{sagar.sakhinana@tcs.com, venkat.runkana@tcs.com} 
}

\begin{document}

\maketitle

\vspace{-2mm}
\begin{abstract}
\vspace{-5mm}
Molecule design is a multifaceted approach that leverages computational methods and experiments to optimize molecular properties, fast-tracking new drug discoveries, innovative material development, and more efficient chemical processes. Recently, text-based molecule design has emerged, inspired by next-generation AI tasks analogous to foundational vision-language models. Our study explores the use of knowledge-augmented prompting of large language models (LLMs) for the zero-shot text-conditional de novo molecular generation task. Our approach uses task-specific instructions and a few demonstrations to address distributional shift challenges when constructing augmented prompts for querying LLMs to generate molecules consistent with technical descriptions. Our framework proves effective, outperforming state-of-the-art (SOTA) baseline models on benchmark datasets.
\end{abstract}

\vspace{-8mm}
\section{Introduction}
\begin{figure}[ht!]
\centering
\resizebox{0.90\linewidth}{!}{ 
\hspace*{-0mm}\includegraphics[keepaspectratio,height=5.0cm,trim=0.0cm 3.0cm 0cm 2.0cm,clip]{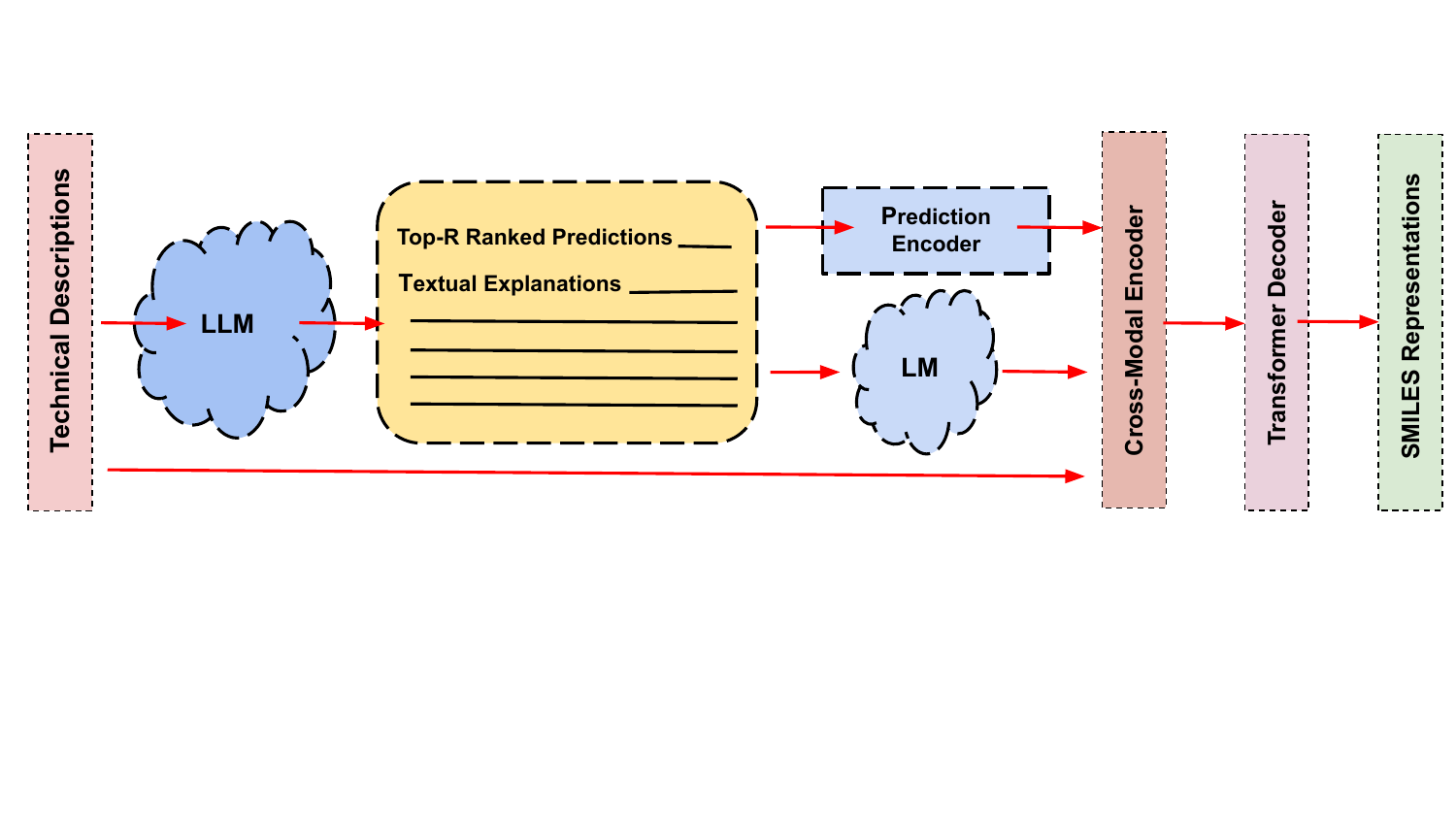} 
}
\vspace{-14mm}
\caption{Overview of the \texttt{FrontierX: LLM-MG} framework. We construct knowledge-augmented prompts using task-specific instructions and a few demonstrations (input-output pairs) based on the downstream task. The augmented prompt queries LLMs to generate the top-$R$ predictions of the SMILES representations and produces textual explanations as justifications for its predictions. We fine-tune small-scale pre-trained language models (LMs) on the generated explanations for domain-specific customization to obtain context-aware token embeddings. We utilize a weighted-sum pooling attention mechanism for task-specific adaptation to compute contextualized text-level embeddings. In parallel, we transform the LLMs' top-$R$ predictions to compute prediction embeddings. The cross-modal encoder, modeled by a hierarchical multi-head attention mechanism, computes the unified embeddings by integrating the mono-domain text-level embeddings (both the original text and explanatory text) and prediction embeddings. Finally, the transformer decoder generates the chemical SMILES representations. We do not repurpose LLMs by fine-tuning with labeled data for domain customization. Instead, we access LLMs via LMaaS\cite{sun2022black} using text-based API interaction.}
\label{fig:figure1}
\vspace{-1mm}
\end{figure}

\vspace{-4mm}
Molecule design is an interdisciplinary approach that involves identifying a target molecule or property to enhance, such as a drug with increased efficacy or a material with superior characteristics. Advancements in science and technology have accelerated the discovery and development of novel drugs, advanced materials, and innovative chemical processes. This iterative process begins with (a) identifying a target molecule or property to improve, followed by (b) employing computational methods to explore the vast chemical space and optimize potential candidate structure and composition. The cycle continues with (c) synthesizing and testing promising candidates in the laboratory until the desired characteristics are achieved. The transformer architecture\cite{vaswani2017attention} has revolutionized various fields in computer science, including language understanding\cite{devlin2018bert}, text generation\cite{radford2019language, brown2020language}, image understanding\cite{dosovitskiy2020image}, and multi-modal generation\cite{ramesh2022hierarchical, saharia2022photorealistic}. Utilizing this architecture to scale language models has established itself as a universal approach for enhancing generalization performance. In recent times, the emergence of foundational Large Language Models (LLMs)\cite{brown2020language, chowdhery2022palm, touvron2023llama}, which are built upon transformer architectures, has significantly revolutionized performance in various natural language processing tasks by enabling enhanced linguistic comprehension and logical reasoning abilities. Different learning strategies such as Zero-Shot Chain of Thought (Zero-shot-CoT\cite{wei2022chain}) and Few-Shot (In-Context) Learning (Few-shot-ICL\cite{rubin2021learning, dong2022survey}) are utilized to leverage the emerging abilities of general-purpose LLMs for a wide variety of specialized tasks across various domains. The former employs task-specific instructions without relying on downstream task-based demonstrations, utilizing the inherent knowledge that the language model acquired during training to generate outputs. In contrast, Few-shot-ICL supplements instructions with a handful of demonstrations, presented as input-output pairs, to foster contextual understanding and facilitate task-specific adaptation, thereby generating relevant output. Recently, there has been a surge in the evolution of generative AI, such as ``DALL·E"\cite{ramesh2022hierarchical, ramesh2021zero} from OpenAI ---  a text-to-image diffusion model that can generate realistic images from text descriptions, and ``Make-A-Video"\cite{singer2022make} from Meta AI ---  a text-to-video diffusion model that generates realistic, engaging, and creative videos from text, among others. Inspired by recent developments in next-generation AI, ``Text-Based Molecule Design"\cite{edwards2022translation} (also known as \texttt{text2mol}) represents a novel cross-domain task in chemistry that involves generating chemical SMILES representations from the corresponding technical descriptions of molecules expressed in natural language. Unlike traditional methods of de novo molecule generation, the \texttt{text2mol} task extracts information from technical descriptions of molecules, identifying aspects such as the specified structure, properties and functional groups, to generate chemical SMILES representations with desired characteristics. Existing models\cite{edwards2022translation, guo2023indeed} in the literature for the \texttt{text2mol} task face challenges in achieving optimal performance and utility, particularly in scenarios where data is scarce and unbalanced. LLMs like ChatGPT\cite{brown2020language}, while proficient in linguistic comprehension, are black-box in nature, resource-intensive, and lack interpretability. Smaller language models(LMs) like BERT\cite{devlin2018bert}, although flexible and interpretable, may lag in reasoning and generalization, resulting in less coherent and contextually relevant responses compared to LLMs. Navigating these challenges requires a delicate balance between performance, efficiency, and interpretability. Our study introduces a novel approach for the \texttt{text2mol} task by combining the strengths of both LLMs and small-scale LMs. LLMs predict a ranked list of chemical SMILES representations while providing explanations as justifications for these predictions, conditioned on the input prompt. These textual explanations, in conjunction with original technical descriptions of molecules, are used to fine-tune small-scale LMs to obtain context-aware token embeddings that capture the essence of both the generated explanations and original text, respectively. Concurrently, the top-ranked predictions generated by LLMs are transformed to obtain prediction embeddings.  By integrating these various embeddings through a hierarchical multi-head attention mechanism, the framework inputs a unified cross-modal embedding into a transformer decoder to generate chemical SMILES representations that align with original technical descriptions. In this study, we explored the use of knowledge-augmented LLM prompting for zero-shot text-conditional molecule generation, a sequence-to-sequence cross-domain task. We present a powerful new tool, \texttt{FrontierX: LLM-MG}, where the goal is to task LLMs with a knowledge-infused prompt that consists of a few demonstrations(input-output pairs) for the text2mol task, along with task-specific instructions, where the output is chemical SMILES representations of the corresponding query technical descriptions. Our experiments on benchmark datasets provide empirical evidence supporting the framework's effectiveness in text-based molecule design tasks. The workflow of the proposed approach is illustrated in Figure \ref{fig:figure1}.

\vspace{-4mm}

\vspace{-7mm}
\section{Proposed Method}
\vspace{-4mm}
The Large Language Models (LLMs), such as ChatGPT\cite{brown2020language}, Meta's LLaMA\cite{touvron2023llama} — that have been pre-trained on large text corpora and operate based on a ``prompt and predict" approach (utilizing natural language prompts to generate the subsequent contextual word or phrase, aligning with human-like responses) — have revolutionized language modeling with their proficiency in linguistic comprehension and advanced logical reasoning abilities, providing improved performance on general-purpose NLP tasks. While LLMs are inherently black box in nature, they possess remarkable capabilities. However, their widespread adoption for applications in various downstream tasks is hindered by the unavailability of logits or token embeddings, which limits explainability. Additionally, they require significant computational resources for fine-tuning on labeled data for task-specific adaptation or for repurposing for domain-customization. In contrast, the small-scale language models (LMs), such as BERT\cite{devlin2018bert} and DeBERTa\cite{he2020deberta}, following a ``pre-train, fine-tune" approach, offer more affordable flexibility for fine-tuning with minimal labeled data and provide access to logits or token embeddings, aiding interpretability. While smaller LMs can learn complex patterns, they often fall short in reasoning and generalization abilities compared to LLMs, which generate more coherent and contextually relevant responses. To alleviate resource constraints, Language Modeling as a Service (LMaaS\cite{sun2022black}) offers access to LLMs through text-based API interactions, while remaining scalable and cost-effective. However, the potential of LLMs for text-conditional de novo molecular generation tasks remains largely underexplored. Our proposed approach for the \texttt{text2mol} task leverages LLMs by utilizing: (a) their predictive ability to provide a top-$R$ ranked list of chemical SMILES representations; and (b) their generative ability to offer auxiliary explanations as justifications for their predictions by conditioning on the augmented prompt. Furthermore, we fine-tune two different small-scale LMs using (a) generated explanations from LLMs and (b) input technical descriptions of molecules to compute their respective contextualized token embeddings --- which capture semantic coherence and contextual relevance for text-to-molecule generation tasks. We utilize weighted attention mechanism to compute both original and explanatory text-level embeddings from their respective context-aware token embeddings. In addition, we transform the LLMs' top-$R$ predictions of chemical SMILES representations into predictive embeddings. We use a hierarchical multi-head attention mechanism to integrate various embeddings into unified cross-modal embedding for input into a transformer decoder, generating the chemical SMILES representation.

\vspace{-2mm}
\paragraph{Evaluation LLMs \& LMs:}
\vspace{-2mm}
In this work, we evaluated three popular LLMs: text-davinci-003\footnote{https://platform.openai.com/docs/models/gpt-3-5}, ChatGPT\footnote{https://chat.openai.com/chat}, and Google BARD\footnote{https://bard.google.com}, in order to thoroughly compare their distinct strengths. text-davinci-003 was the earliest LLM released by OpenAI and was tailored for a broad spectrum of linguistic tasks. GPT-3.5-turbo is a substantial improvement over the GPT-3 base models, demonstrating remarkable performance on a wide range of linguistic tasks while also being cost-effective. Google BARD\cite{anil2023palm} stands out due to its extraordinary scale, complexity, and an impressively extensive vocabulary compared to the GPT-3.5 models. In addition to these, our study also incorporates a pre-trained smaller LM, DeBERTa\footnote{For more information on DeBERTa, please refer to \url{https://huggingface.co/docs/transformers/index} \cite{he2020deberta}}, which is an improved version of the BERT\cite{devlin2018bert} architecture. Table \ref{model} presents a comprehensive summary of the technical specifications of these language models.

\vspace{-3mm}
\begin{table}[ht!]
	\centering
	\small
	\caption{Technical details of LLMs and LMs. \emph{Enterprise} refers to the organization that developed the language models. \emph{Cost} denotes the expenses associated with using 1K tokens. \emph{Last Update Date} indicates that the LLM's knowledge base is limited to information available up to that specific date.}
	\vspace{-2mm}
	\hspace{-5mm}\begin{tabular}{c|c|c|c|c}
            \toprule
            \textbf{Model} &\textbf{Enterprise} &\textbf{Cost}  &\textbf{Last Update Date} &\textbf{Vocabulary Size}\\
            \hline
            text-davinci-003 &Open-AI &0.02\$ &Sep. 2021 &175B  \\
            ChatGPT &Open-AI &0.002\$ &Jun. 2021 & 175B \\
            BARD & Google &Free & Undisclosed & 1,560B \\
            \hline
            DeBERTa & Hugging Face & Free & N/A & 50M \\
            \hline
	\end{tabular}
	\vspace{-0.2cm}
	\label{model}
\end{table}

\vspace{-3mm}
\paragraph{Knowledge-Augmented Prompts:} 
In our work, we offer essential context and task-specific instructions by using input natural language descriptions of the target molecule to prompt LLMs in a zero-shot setting to generate corresponding chemical SMILES representations. In this scenario, the primary task-specific instructions involve the translation of these descriptions into chemical SMILES representations. We create an augmented prompt that incorporates both the task-specific instructions and a few demonstrations. These demonstrations, which establish the context, are grounded in the downstream \texttt{text2mol} task and comprise input-output pairs (i.e., technical descriptions and their corresponding chemical SMILES representations). This approach facilitates knowledge-augmented prompting of the LLMs for zero-shot text-to-molecule generation tasks. The construction of an augmented prompt involves sampling text-molecule pairs from the training data that are relevant to the target molecule descriptions. We then prepend these pairs to the task-specific instructions to form an augmented prompt, which is used to query the LLMs in a zero-shot setting for the generation of chemical SMILES representations. To evaluate the impact of the quality and quantity of sampled text-molecule pairs on the performance of text-conditional de novo molecule generation tasks, we employ two different sampling strategies. The quality of these pairs is determined by the sampling methods used to identify pairs similar to the target molecule descriptions. We navigate through the training dataset using two semantic search-retrieval methodologies --- \texttt{random} and \texttt{scaffold} --- to sample text-molecule pairs relevant to the target molecule descriptions. The random approach involves arbitrarily sampling \textit{K} text-molecule pairs from the training dataset. In contrast, the scaffold technique employs semantic similarity methods, specifically \texttt{text-embedding-ada-002} from OpenAI\footnote{https://platform.openai.com/docs/guides/embeddings}, to evaluate the similarity between the molecular textual descriptions in the training dataset and the target molecule descriptions. It then selects the top-\textit{K} most relevant text-molecule pairs, where the hyperparameter \textit{K} is set using a random search technique. We employ the different sampling strategies to analyze the effectiveness of augmenting prompts with relevant text-molecule pairs in language-conditioned molecule generation tasks. In short, unlike traditional supervised learning, LLMs (a) predict the chemical SMILES representations and (b) generate textual explanations for their predictions, utilizing the inherent knowledge embedded within the language model's parameters, all conditioned on the augmented prompt, without needing any parameter updates.
  
\vspace{-4mm} 
\paragraph{Querying LLMs:} 
We access LLMs with LMaaS\cite{sun2022black} platforms via text-based API interaction, necessitating solely text-based input and output. We create a customized zero-shot prompt template to query LLMs to translate textual descriptions into chemical SMILES representations. The LLMs' response serves the dual purpose of (a) providing detailed textual explanations for the underlying rationale, (reasoning or logic), behind the predictions and (b) generating a list of the top-$R$ ranked chemical SMILES representations. Subsequently, we fine-tune smaller downstream LMs using the generated auxiliary explanations. The custom augmented prompt format is as follows:

\vspace{-2mm}
\begin{tcolorbox}
\centering
\vspace{-2mm}
Below are the textual descriptions -- chemical SMILES representation pairs. Generate the chemical SMILES representation for the textual description provided below.
\vspace{-2mm}
\end{tcolorbox}

\vspace{-1mm}
Querying LLMs (a) predicts the top-$R$ ranked chemical SMILES representations and (b) provides auxiliary explanations as logical justifications for its predictions.

\vspace{-2mm}
\begin{tcolorbox}
\vspace{-4mm}
\centering
(\textbf{LLMs Response}) [top-$R$ ranked predictions --- Auxiliary Explanations]
\vspace{-2mm}
\end{tcolorbox}

\vspace{-2mm}
In the next section, we will discuss the use of auxiliary explanations and original textual descriptions for fine-tuning various downstream smaller LMs for domain customization. Later, we will transform the LLMs top-$R$ predictions of chemical SMILES representations into predictive embeddings.

\vspace{-4mm}
\paragraph{Fine-tuning LMs for Domain-Specific Customization:} Our novel approach leverages the integration of a smaller language model (LM) to extract relevant information from the original molecular textual descriptions and auxiliary explanations generated by LLMs, thereby aiding downstream tasks. The intermediary LM serves as a bridge between the LLM and the downstream layers that generate chemical SMILES representations. To elucidate further, we fine-tune pre-trained LMs, denoted as $\textrm{LM}_{\textrm{exp}}$ and $\textrm{LM}_{\textrm{org}}$, to compute context-aware token embeddings by passing the text sequences generated by LLMs (referred to as $\mathcal{S}_{\textrm{exp}}$) and original textual descriptions (referred to as $\mathcal{S}_{\textrm{org}}$) through the $\textrm{LM}_{\textrm{exp}}$ and $\textrm{LM}_{\textrm{org}}$ models, respectively, as described below:

\vspace{-2.5mm}
\resizebox{0.93\linewidth}{!}{
\hspace{0mm}\begin{minipage}{\linewidth}
\begin{equation}
h_{\textrm{exp}} = \textrm{LM}_\textrm{exp}(\mathcal{S}_{\textrm{exp}}) \in \mathbb{R}^{(m \times d)}; \quad h_{\textrm{org}} = \textrm{LM}_\textrm{org}(\mathcal{S}_{\textrm{org}}) \in \mathbb{R}^{(n\times d)}
\end{equation}
\end{minipage}
} 

\vspace{-1mm}
where both contextualized embeddings $h_{\textrm{exp}}$ and $h_{\textrm{org}}$ capture not only the contextual information of the tokens but also encapsulate the semantic relationships among tokens within their respective textual content. Here, $\text{m}$ and $\text{n}$ represent the number of tokens in $\mathcal{S}_{\textrm{exp}}$ and $\mathcal{S}_{\textrm{org}}$, while $d$ represents the token embedding dimension. We employ a softmax attention mechanism to compute a weighted sum of the contextualized token embeddings, encoding the auxiliary explanations and original textual descriptions into single fixed-length vectors or embeddings denoted as $y_{\textrm{exp}}$ and $y_{\textrm{org}}$ and computed as follows,

\vspace{-5mm}
\resizebox{0.97\linewidth}{!}{
\hspace{0mm}\begin{minipage}{\linewidth}
\begin{equation}
\alpha_i = \mbox{softmax}(q_i); \hspace{2mm} q_i = \mathbf{u}^Th^{(i)}_{\textrm{exp}} \hspace{2mm} || \hspace{2mm}\beta_i = \mbox{softmax}(r_i); \hspace{2mm} r_i = \mathbf{v}^Th^{(i)}_{\textrm{org}}
\end{equation}
\end{minipage}
}

\vspace{-2mm}
\resizebox{0.935\linewidth}{!}{
\hspace{0mm}\begin{minipage}{\linewidth}
\begin{equation}
y_{\text{exp}} = \sum_{i=0}^\text{m}{\alpha_i h^{(i)}_{\textrm{exp}}} \in \mathbb{R}^{(d)}; \hspace{2mm} y_{\text{org}} = \sum_{i=0}^\text{n}{\beta_i h^{(i)}_{\textrm{org}}} \in \mathbb{R}^{(d)}
\end{equation}
\end{minipage}
} 

\vspace{-1mm}
where $\mathbf{u}$ and $\mathbf{v}$ are differentiable vectors. The explanatory text-level embedding, represented as $y_{\text{exp}}$, encapsulates domain-specific knowledge retrieved from foundational LLMs to support its predictions. The original text-level embedding, denoted as $y_{\text{org}}$, captures the overall context and semantics within the original textual descriptions by extracting the most pertinent and task-relevant information.

\vspace{-4mm}
\paragraph{LLMs Prediction Embeddings:} As mentioned earlier, the LLMs not only provide the auxiliary textual explanations but also predict the top-$R$ ranked chemical SMILES representations list, which can be informative. For each target molecule in the \texttt{text2mol} task, the top-$R$ predictions are converted into one-hot encoded vectors $p_{i,1}, \ldots, p_{i,R} \in \mathbb{R}^C$, where $C$ represents the total number of elements in the SMILES vocabulary, encompassing a wide range of characters and symbols used to represent chemical structures. These vectors are subsequently concatenated into a single $RC$-dimensional vector, and finally, they undergo linear encoding into a fixed-length prediction embedding $y_{\text{pred}} \in \mathbb{R}^{d}$, encapsulating the top-$R$ predictions from the LLMs.

\vspace{-4mm}
\paragraph{Cross-modal Attention Layer } We compute the cross-modal embedding, denoted as $y_{\text{cross}}$, using a hierarchical multi-head attention mechanism that integrates the original text-level embedding $y_{\text{org}}$, the explanatory text-level embedding $y_{\text{exp}}$, and the prediction embedding $y_{\text{pred}}$. This mechanism provides a robust framework for integrating diverse information encapsulated from different modalities, addressing several key aspects critical to the performance of cross-modal learning tasks. It involves hierarchical implementation of multi-head attention mechanisms. We employ two layers, each focusing on different aspects of the input embeddings, enabling more complex interactions and potentially leading to more scalable and efficient models. In the initial layer, we apply a multi-head attention mechanism to the mono-domain embeddings, specifically the original text-level embedding $y_{\text{org}}$ and the explanatory text-level embedding $y_{\text{exp}}$, to obtain unified mono-domain embeddings denoted as $y_\text{uni}$. In the subsequent layer, we utilize the multi-head attention mechanism on the cross-domain embeddings, comprising the prediction embedding $y_{\text{pred}}$ and the unified mono-domain embeddings $y_\text{uni}$, to compute the cross-modal embeddings denoted as $y_\text{cross}$. For the first layer, we  compute the Query, Key, Value projections for the original text-level embedding $y_{\text{org}}$ for each head h as follows:

\vspace{-7mm}
\resizebox{0.945\linewidth}{!}{
\hspace{0mm}\begin{minipage}{\linewidth}
\begin{align}
Q^h_{\text{org}} &= y_{\text{org}} W^h_{Q_\text{org}}; K^h_{\text{org}} = y_{\text{org}} W^h_{K_\text{org}}; V^h_{\text{org}} = y_{\text{org}} W^h_{V_\text{org}}
\end{align}
\end{minipage}
} 

\vspace{-1mm}
Similarly, the Query, Key, Value projections for explanation text-level embedding $y_{\text{exp}}$ for each head $h$ as follows:

\vspace{-7mm}
\resizebox{0.945\linewidth}{!}{
\hspace{0mm}\begin{minipage}{\linewidth}
\begin{align}
Q^h_{\text{exp}} &= y_{\text{exp}} W^h_{Q_{\text{exp}}}; K^h_{\text{exp}} = y_{\text{exp}} W^h_{K_{\text{exp}}}; V^h_{\text{exp}} = y_{\text{exp}} W^h_{V_{\text{exp}}}
\end{align}
\end{minipage}
} 

\vspace{-1mm}
We concatenate the keys and values from both original and explanatory text-level embeddings, which provides a powerful way to integrate information from the mono-domain embeddings into a unified, rich representation. 

\vspace{-8mm}
\resizebox{0.945\linewidth}{!}{
\hspace{3mm}\begin{minipage}{\linewidth}
\begin{align}
K^h_{\text{concat}} &= [K^h_{\text{org}}, K^h_{\text{exp}}]; V^h_{\text{concat}} = [V^h_{\text{org}}, V^h_{\text{exp}}]
\end{align}
\end{minipage}
} 

\vspace{-1mm}
We use softmax attention to integrate complementary information from the mono-domain embeddings, focus on contextually relevant information, and semantically align them through an attention mechanism. The softmax function is applied to the keys for each query.

\vspace{-4mm}
\resizebox{0.875\linewidth}{!}{
\hspace{0mm}\begin{minipage}{\linewidth}
\begin{align}
A^h_{\text{uni}} &= \text{Softmax}\left(\frac{(Q^h_{\text{org}} + Q^h_{\text{exp}}) {K^h_{\text{concat}}}^T}{\sqrt{d_h}}\right)
\end{align}
\end{minipage}
} 

\vspace{-2mm}
Each head outputs a new vector representation that highlights the most relevant features in the mono-domain embeddings(both original and explanation text-level), according to the attention mechanism for that specific head, which is tailored to capture specific aspects or relationships within the data.

\vspace{-4mm}
\resizebox{0.945\linewidth}{!}{
\hspace{0mm}\begin{minipage}{\linewidth}
\begin{align}
O^h_{\text{uni}} &= A^h_{\text{uni}} V^h_{\text{concat}}
\end{align}
\end{minipage}
} 

\vspace{-2mm}
Finally, all the head-specific outputs are concatenated and linearly transformed to create the unified mono-domain embedding as follows,

\vspace{-5mm}
\resizebox{0.925\linewidth}{!}{
\hspace{0mm}\begin{minipage}{\linewidth}
\begin{align}
O_{\text{concat}} &= [O^1_{\text{uni}}, O^2_{\text{uni}}, \ldots, O^H_{\text{uni}}] \\
y_{\text{uni}} &= O_{\text{concat}} W_{O_{\text{uni}}}
\end{align}
\end{minipage}
} 

\vspace{-1mm}
where $W^h_{Q_\text{org}}$, $W^h_{K_\text{org}}$, $W^h_{V_\text{org}}$, $W^h_{V_\text{exp}}$,  $W^h_{Q_{\text{exp}}}$,  $W^h_{K_{\text{exp}}}$, $W_{O_{\text{uni}}}$ are the learnable weight matrices. Here, $d_h$ represents the dimensionality of the key/query/value for each head, and $H$ is the number of heads. $y_{\text{uni}}$ denotes the unified mono-domain embeddings. The unified embeddings can learn and integrate complementary, diverse information present in both the $y_{\text{org}}$ and $y_{\text{exp}}$ embeddings. These unified embeddings facilitate semantic alignment among similar features across different embeddings and enable the identification of contextual relevance between distinct yet related $y_{\text{org}}$ and $y_{\text{exp}}$ mono-domain embeddings. The next step involves computing the cross-modal embedding $y_{\text{cross}}$ using a second layer of a multihead attention mechanism that integrates both $y_{\text{pred}}$ and $y_{\text{uni}}$. We  compute the Query, Key, and Value projections for the prediction embedding $y_{\text{pred}}$ for each head h as follows:

\vspace{-4mm}
\resizebox{0.945\linewidth}{!}{
\hspace{0mm}\begin{minipage}{\linewidth}
\begin{align}
Q^h_{\text{pred}} &= y_{\text{pred}} W^h_{Q_\text{pred}}; K^h_{\text{pred}} = y_{\text{pred}} W^h_{K_\text{pred}}; V^h_{\text{pred}} = y_{\text{pred}} W^h_{V_\text{pred}}
\end{align}
\end{minipage}
} 

\vspace{-1mm}
Similarly, we  compute the Query, Key, Value projections for the unified embedding $y_{\text{uni}}$ for each head h as follows:

\vspace{-7mm}
\resizebox{0.945\linewidth}{!}{
\hspace{0mm}\begin{minipage}{\linewidth}
\begin{align}
Q^h_{\text{uni}} &= y_{\text{uni}} W^h_{Q_\text{uni}}; K^h_{\text{uni}} = y_{\text{uni}} W^h_{K_\text{uni}}; V^h_{\text{uni}} = y_{\text{uni}} W^h_{V_\text{uni}}
\end{align}
\end{minipage}
} 

\vspace{-1mm}
We concatenate the keys and values from both the prediction and unified embeddings, thereby facilitating a robust integration of insights from the cross-domain embeddings into a synergized and enriched representation.

\vspace{-7mm}
\resizebox{0.98\linewidth}{!}{
\hspace{10mm}\begin{minipage}{\linewidth}
\begin{align}
K^h_{\text{cross}} &= [K^h_{\text{uni}}, K^h_{\text{pred}}]; V^h_{\text{cross}} = [V^h_{\text{uni}}, V^h_{\text{pred}}]
\end{align}
\end{minipage}
} 

\vspace{-1mm}
We utilize a softmax attention mechanism to merge and align information from different domains, thereby prioritizing contextually relevant information and ensuring semantic alignment. 
The softmax
\newpage
function is applied to the keys for each query, described as follows:

\vspace{-5mm}
\resizebox{0.875\linewidth}{!}{
\hspace{0mm}\begin{minipage}{\linewidth}
\begin{align}
A^h_{\text{cross}} &= \text{Softmax}\left(\frac{(Q^h_{\text{uni}} + Q^h_{\text{pred}}) {K^h_{\text{cross}}}^T}{\sqrt{d_h}}\right)
\end{align}
\end{minipage}
} 

\vspace{-2mm}
In the multi-head attention mechanism, each head processes both embeddings(unified and predictive embeddings) to highlight important patterns, focusing on specific relationships or aspects within the data, enhancing performance in cross-modal learning tasks.

\vspace{-4mm}
\resizebox{0.925\linewidth}{!}{
\hspace{0mm}\begin{minipage}{\linewidth}
\begin{align}
O^h_{\text{cross}} &= A^h_{\text{cross}} V^h_{\text{cross}}
\end{align}
\end{minipage}
} 

\vspace{-1mm}
Finally, all the head-specific outputs are concatenated and linearly transformed to create the final cross-modal embedding as follows,

\vspace{-4mm}
\resizebox{0.935\linewidth}{!}{
\hspace{0mm}\begin{minipage}{\linewidth}
\begin{align}
O_{\text{cross}} &= [O^1_{\text{cross}}, O^2_{\text{cross}}, \ldots, O^H_{\text{cross}}] \\
y_{\text{cross}} &= O_{\text{cross}} W_{O_{\text{cross}}}
\end{align}
\end{minipage}
} 

where $W^h_{Q_\text{uni}}$, $W^h_{K_\text{uni}}$, $W^h_{V_\text{uni}}$, $W^h_{V_\text{pred}}$,  $W^h_{Q_{\text{pred}}}$,  $W^h_{K_{\text{pred}}}$, $W_{O_{\text{cross}}}$ are the learnable weight matrices. $y_{\text{cross}}$ denotes the cross-domain embeddings. Implementing the hierarchical attention mechanism facilitates the structured integration of information from different modalities. This mechanism employs multi-head attention method, using multiple sets of learned weight matrices to emphasize various aspects or relationships within the data. Consequently, this approach has the potential to foster robust and enriched embeddings capable of capturing complex patterns. Additionally, it aids in focusing on contextually pertinent information and achieving semantic alignment across different embeddings, thereby enhancing the capacity to identify and utilize crucial features in the input data.
 
\vspace{-4mm}
\paragraph{Output Layer:}  We then utilize a transformer decoder\cite{vaswani2017attention} to generate chemical SMILES representations character by character, using cross-modal embeddings ($y_{\text{cross}}$) that incorporate global context through the hierarchical multi-head self-attention mechanism. We implement a softmax layer to transform the decoder's output, creating a probability distribution over potential elements for each position in the SMILES strings. For our sequence generation tasks, we minimize the categorical cross-entropy loss to penalize the proposed framework based on the negative log-likelihood of the ground-truth chemical SMILES strings under the predicted probability distribution, thus facilitating the generation of valid molecules. In summary, by integrating multi-modal embeddings, namely $y_{\text{org}}$, $y_{\text{exp}}$, and $y_{\text{pred}}$, our approach enables the concurrent capture of complementary information, ultimately enhancing the overall performance of the framework.

\vspace{-5mm}
\section{Experiments \& Results}
\vspace{-5mm}

\subsection{Datasets \& Baselines}\label{sec:baselines}
\vspace{-3mm}
Our study utilized the ChEBI-20 dataset\cite{edwards2022translation}, a bidirectional text-to-molecule translation dataset comprising 33,010 text description-molecule pairs with a predefined split ratio of 80:10:10 for training, validation, and test sets, respectively. We utilized 26,407 text description-molecule pairs from the training set for demonstrations (input-output mappings) for constructing a knowledge-augmented prompt to query LLMs. We used the MolT5 model\cite{edwards2022translation}, as a predominant baseline, which is an encoder-decoder transformer architecture pretrained on a large unannotated dataset specifically for the \texttt{text2mol} translation task, building upon the foundations of the T5\cite{raffel2020exploring} model. We evaluated the performance of our proposed framework on the \texttt{text2mol} task, comparing it with several variants of the MolT5\cite{edwards2022translation} and T5\cite{raffel2020exploring} models, as well as with general-purpose sequence-to-sequence models, such as the RNN-GRU and Vanilla Transformer models. In addition, various variants of few-shot (ICL) prompting of GPT-based models --- reflecting the fact that this technique uses few-shot learning  to prompt off-the-shelf GPT-based models to perform molecular property prediction for new, unseen molecules --- referred to as baselines, are evaluated for comparison with our proposed framework. The configurations include different variants of the GPT-4 model, namely, (a) the zero-shot approach, (b) the scaffold sampling technique with $K$=10 or $K$=5, and (c) the random sampling technique with $K$=10 for constructing augmented prompts. In addition, we use GPT-3.5 and davinci-003 models, both employing the scaffold sampling technique with $K$=10 to construct knowledge-augmented prompts. For more details and information on the baselines, please refer to the earlier works\cite{guo2023indeed, edwards2022translation}.

\vspace{-4mm}
\subsection{Evaluation Metrics}
\vspace{-3mm}
To comprehensively evaluate the quality and similarity of the generated chemical SMILES representations compared to the ground-truth SMILES representations, we employed a range of distinct evaluation metrics, categorized into three types. These metrics include (a) chemical similarity measures, such as the FTS (Fingerprint Tanimoto Similarity) \cite{tanimoto1958elementary} and the FCD (Fréchet ChemNet Distance) \cite{preuer2018frechet}, as well as (b) natural language processing metrics like the BLEU (Bilingual Evaluation Understudy) score, Exact Match \cite{edwards2022translation}, and Levenshtein distance \cite{miller2009levenshtein}. In addition, (c) we utilized the RDKit library \cite{rdkit} to validate the generated molecules. We delineate the metrics as follows: (a) We employ the FTS\cite{tanimoto1958elementary} metric to gauge the chemical similarity between the ground-truth and generated chemical compounds represented as SMILES strings(notation to represent chemical structures as text) by comparing their MACCS, RDK, and Morgan fingerprints\cite{SequenceMatcher, rdkit, dash2023evaluation}. (b) In addition, we utilize the FCD metric\cite{preuer2018frechet}, which leverages latent information from a pretrained model\cite{preuer2018frechet} to predict molecular activity\cite{edwards2022translation}. The FCD is calculated by measuring the distance between the mean embeddings of two sets of chemical SMILES strings (generated and ground-truth) in the latent space of the pretrained model. A lower FCD score indicates a greater similarity between the corresponding
two sets of molecules. (c) We also apply natural language processing metrics to evaluate the quality of the chemical SMILES strings generated by our framework. These metrics encompass the following: (i) BLEU  --- this measures the similarity between two text strings, with a higher BLEU score denoting better similarity. (ii) Exact Match\cite{edwards2022translation} --- this quantifies the percentage of generated chemical SMILES strings that are identical to the ground-truth strings. (iii) Levenshtein distance\cite{miller2009levenshtein} --- this calculates the minimum number of single-character edits required to modify the generated chemical SMILES strings to match the ground-truth strings, with a lower value indicating closer similarity. By utilizing these diverse metrics, we attain a nuanced understanding of the efficacy of our text-conditional de novo molecule generation framework. Higher FTS scores and lower FCD scores signify better chemical similarity and closer molecular activity resemblance, respectively. In the context of matching chemical SMILES strings from the perspective of natural language processing, higher BLEU and Exact Match scores are preferred to achieve better alignment with the ground-truth SMILES strings, while a lower Levenshtein distance indicates fewer required edits, denoting superior similarity. The RDKit library assists in verifying the validity of the generated molecules, with a higher proportion indicating successful generation\cite{rdkit}.

\vspace{-5mm}
\subsection{Experimental Setup}
\vspace{-3mm}
We used the ChEBI-20 dataset\cite{edwards2021text2mol} with an 80:10:10 split: 80$\%$ for training, 10$\%$ for validation, and 10$\%$ for testing. The training set was utilized to update the learnable parameters, the validation set to select optimal hyperparameters, and the test set to evaluate generalization performance of the proposed framework. Our scalable and efficient framework offers a unified solution for integrating LLMs and LMs. We configured the hyperparameters of our framework with a batch size of 32, trained it for 100 epochs, and a hidden or embedding dimension($d$) of 128. Additional hyperparameters include the number of attention heads (H) set to 4, and the dimensionality of Key/Query/Value ($d_{h}$) is 32. To optimize the training process, we utilized the Adam optimizer \cite{kingma2014adam}, initially setting the learning rate to $1e^{-3}$. Additionally, we incorporated a learning rate decay scheduler, which reduced the learning rate by half whenever the validation loss did not improve for 10 consecutive epochs. Furthermore, we applied early stopping to prevent overfitting on the validation data. We evaluated our approach using the following LLMs: GPT-4.0, GPT-3.5-turbo, GPT-3.0-text-davinci-003, and Google Bard. In our approach, we chose not to fine-tune hyperparameters individually for each LLM, opting instead to maintain consistent settings across all language models. This strategy simplifies experimentation, ensures uniform conditions, facilitates result comparison, and promotes consistency. Moreover, it underscores the versatility of our framework, which can be used with any off-the-shelf LLM without the need for computationally expensive hyperparameter tuning. We used the Scaffold technique with $K$=16 to sample demonstrations (input-output mappings) from the training data to construct augmented prompts for querying LLMs in few-shot settings. In addition, we query LLMs to generate the top-$R$ ranked chemical SMILES strings predictions list and set the hyperparameter $R$ as 4. To maximize computational resource utilization, we harnessed eight V100 GPUs, each equipped with 8 GB of GPU memory, for training deep learning models built upon the PyTorch framework. Considering the context length limitations imposed by LLMs, which restrict the maximum sequence length that a typical LLM can process at a time to 4096 tokens, we implemented strategies to mitigate the high computational costs associated with prompting LLMs. This approach included running each experiment twice and reporting the average results. Our approach prioritizes both resource optimization and accuracy, aiming to achieve the best possible outcomes while minimizing the computational footprint. Our evaluation incorporated several metrics, and we present the results for the test datasets and compare the performance against well-known baselines.

\vspace{-5mm}
\subsection{Results}
\vspace{-3mm}
The experimental results of the proposed framework and the baseline models performance on the \texttt{text2mol} task are presented in Tables \ref{text_based_moldesign_results1} and \ref{text_based_moldesign_results2}. The results of the baseline models are reported from earlier studies \cite{guo2023indeed, edwards2022translation}. The results undeniably demonstrate the superior performance of the $\textbf{FrontierX: LLM-MG}$ framework, especially when combined with the GPT-4 backbone and employing the Scaffold technique with $K$ set to 16. This optimal combination excels in generating accurate molecular structures that closely resemble the ground truth, surpassing all baseline models across

\newpage
various evaluation metrics.

\vspace{-2mm}
\begin{table}[!ht]
\centering
\caption{The table presents a performance comparison of the proposed framework and the baselines on the \texttt{text2mol} task. The top-performing model is highlighted in bold. The baseline results are reported from previous work\cite{guo2023indeed}. We leveraged the Scaffold technique, setting $K$ to 16, to sample demonstrations and construct augmented prompts for in-context learning in all experiments involving \texttt{FrontierX: LLM-MG} with various off-the-shelf LLMs.}
\vspace{-1mm}
\begin{adjustbox}{width=1.225\columnwidth,center}
\hspace{-5mm}\begin{tabular}{@{}|c|cccccccc|@{}}
\toprule
\toprule
\textbf{Method} & \textbf{BLEU ($\uparrow$)}                       & \textbf{Exact ($\uparrow$)}                      & \textbf{Levenshtein ($\downarrow$)}                 & \textbf{Validity ($\uparrow$)}                   & \textbf{MACCS FTS ($\uparrow$)}                  & \textbf{RDK FTS ($\uparrow$)}                    & \textbf{Morgan FTS ($\uparrow$)}                 & \textbf{FCD ($\downarrow$)}                        \\ \midrule
\begin{tabular}[c]{@{}l@{}} 
Ground-Truth \\ \end{tabular}            & 1.0          & 1.0          & 0.0          & 1.0          & 1.0          & 1.0          & 1.0          & 0         \\ \midrule
\begin{tabular}[c]{@{}l@{}} GPT-4 \\ (zero-shot)\end{tabular}            & 0.490$\pm$0.017          & 0.046$\pm$0.009          & 47.418$\pm$1.668          & 0.758$\pm$0.015          & 0.733$\pm$0.020          & 0.514$\pm$0.021          & 0.432$\pm$0.014          & 11.913$\pm$0.97         \\ \midrule
\begin{tabular}[c]{@{}l@{}}GPT-4 \\ (Scaffold, $k$=10)\end{tabular}       & 0.816$\pm$0.004 & 0.174$\pm$0.029 & 21.160$\pm$0.600 & 0.888$\pm$0.023          & 0.867$\pm$0.005 & 0.738$\pm$0.010          & 0.672$\pm$0.013 & 6.224$\pm$0.44          \\ \midrule
\begin{tabular}[c]{@{}l@{}}GPT-4 \\ (Scaffold, $k$=5)\end{tabular}        & 0.815$\pm$0.011          & 0.164$\pm$0.018          & 21.862$\pm$1.768          & 0.874$\pm$0.030          & 0.865$\pm$0.015          & 0.741$\pm$0.023 & 0.670$\pm$0.028          &  5.843$\pm$0.515 \\ \midrule
\begin{tabular}[c]{@{}l@{}}GPT-4 \\ (Random, $k$=10)\end{tabular}         & 0.602$\pm$0.016          & 0.060$\pm$0.007          & 42.390$\pm$1.008          & 0.770$\pm$0.030          & 0.762$\pm$0.013          & 0.548$\pm$0.017          & 0.475$\pm$0.015          & 10.594$\pm$0.41         \\ \midrule
\begin{tabular}[c]{@{}l@{}}GPT-3.5 \\ (Scaffold, $k$=10)\end{tabular}     & 0.479$\pm$0.156          & 0.094$\pm$0.011          & 82.008$\pm$40.354         & 0.854$\pm$0.059          & 0.833$\pm$0.006          & 0.686$\pm$0.016          & 0.585$\pm$0.013          & 8.341$\pm$0.607          \\ \midrule
\begin{tabular}[c]{@{}l@{}}Davinci-003 \\ (Scaffold, $k$=10)\end{tabular} & 0.741$\pm$0.011          & 0.100$\pm$0.010          & 25.648$\pm$2.186          & 0.936$\pm$0.009 & 0.783$\pm$0.014          & 0.648$\pm$0.004          & 0.560$\pm$0.010          & 8.335$\pm$0.310          \\ \midrule 
\begin{tabular}[c]{@{}l@{}}\textbf{FrontierX-} \\ \textbf{W/GPT-4}\end{tabular}  & \textbf{0.937$\pm$0.023} & \textbf{0.641 $\pm$0.089}  & \textbf{6.946 $\pm$0.045} & \textbf{0.975 $\pm$0.013} &  \textbf{0.947 $\pm$0.029} & \textbf{0.838$\pm$0.006} & \textbf{0.845$\pm$0.012} & \textbf{0.796$\pm$0.003} \\ \midrule
\begin{tabular}[c]{@{}l@{}}\textbf{FrontierX- } \\ \textbf{W/GPT-3.5-turbo}\end{tabular} & 0.893$\pm$0.051 & 0.567$\pm$0.112 & 11.431 $\pm$0.086 & 0.939 $\pm$0.027 & 0.914 $\pm$0.015 & 0.785$\pm$0.012 & 0.809$\pm$0.016 &  0.968$\pm$0.004 \\ \midrule
\begin{tabular}[c]{@{}l@{}}\textbf{FrontierX- } \\  \textbf{W/GPT-3.0}\end{tabular} & 0.885$\pm$0.084 &  0.536$\pm$0.173 & 13.867 $\pm$0.137 & 0.914 $\pm$0.041 & 0.896 $\pm$0.034 & 0.774$\pm$0.028 & 0.823$\pm$0.034 &  1.025$\pm$0.007\\ \midrule
\begin{tabular}[c]{@{}l@{}}\textbf{FrontierX-} \\ \textbf{W/Google Bard}\end{tabular} & 0.749$\pm$0.109 & 0.426$\pm$0.138 & 16.729 $\pm$0.145 & 0.855 $\pm$0.017 & 0.863 $\pm$0.018 & 0.692$\pm$0.037 & 0.727$\pm$0.045 & 2.345$\pm$0.012\\ \bottomrule
\bottomrule
\end{tabular}
\end{adjustbox}
\label{text_based_moldesign_results1}
\vspace{-2mm}
\end{table}

\vspace{-2mm}
\begin{table}[!ht]
\centering
\caption{The table illustrates the performance of both the proposed framework and the baseline models in the \texttt{text2mol} task. We have highlighted the top-performing model using bold text. Baseline results are reported from earlier research\cite{edwards2022translation}. For all \texttt{FrontierX: LLM-MG} experiments with various off-the-shelf LLMs, we employed the Scaffold technique, setting $K$ to 16 to sample demonstrations for constructing augmented prompts for in-context learning.}
\vspace{-1mm}
\begin{adjustbox}{width=1.225\columnwidth,center}
\hspace{-5mm}\begin{tabular}{@{}|c|cccccccc|@{}}
\toprule
\toprule
\textbf{Method} & \textbf{BLEU ($\uparrow$)}                       & \textbf{Exact ($\uparrow$)}                      & \textbf{Levenshtein ($\downarrow$)}                                    & \textbf{MACCS FTS ($\uparrow$)}                  & \textbf{RDK FTS ($\uparrow$)}                    & \textbf{Morgan FTS ($\uparrow$)}                 & \textbf{FCD ($\downarrow$)}                        & \textbf{Validity ($\uparrow$)}\\ \midrule
\begin{tabular}[c]{@{}l@{}} 
Ground-Truth \\ \end{tabular}            & 1.0          & 1.0          & 0.0          & 1.0          & 1.0          & 1.0          & 0.0          & 1.0         \\ \midrule
RNN-GRU & 0.652 & 0.005 & 38.09 & 0.591 & 0.400 & 0.362 & 4.55 &  0.542 \\ \midrule
Transformer & 0.499 & 0.000 & 57.66 & 0.480 & 0.320 & 0.217 & 11.32 & 0.906 \\ \midrule
T5-Small & 0.741 & 0.064 & 27.703 & 0.704 & 0.578  & 0.525  & 2.89  & 0.608 \\ \midrule
MolT5-Small & 0.755 & 0.079 & 25.988 & 0.703 & 0.568 & 0.517 & 2.49 & 0.721 \\  \midrule
T5-Base & 0.762 & 0.069 & 24.950 & 0.731 &  0.605 & 0.545 & 2.48 & 0.660  \\ \midrule
MolT5-Base & 0.769 & 0.081 & 24.458 & 0.721 &  0.588 & 0.529 & 2.18 & 0.772 \\ \midrule
T5-Large & 0.854 & 0.279 & 16.721 & 0.823 &  0.731 & 0.670 & 1.22 & 0.902 \\ \midrule
MolT5-Large & 0.854 & 0.311 & 16.071 & 0.834 & 0.746 & 0.684 & 1.20 & 0.905 \\ \midrule 
\begin{tabular}[c]{@{}l@{}}\textbf{FrontierX-} \\ \textbf{W/GPT-4}\end{tabular}  & \textbf{0.937$\pm$0.023} & \textbf{0.641 $\pm$0.089}  & \textbf{6.946 $\pm$0.045} &  \textbf{0.947 $\pm$0.029} & \textbf{0.838$\pm$0.006} & \textbf{0.845$\pm$0.012} & \textbf{0.796$\pm$0.003} & \textbf{0.975 $\pm$0.013}\\ \midrule
\begin{tabular}[c]{@{}l@{}}\textbf{FrontierX-} \\ \textbf{W/GPT-3.5-turbo}\end{tabular} & 0.893$\pm$0.051 & 0.567$\pm$0.112 & 11.431 $\pm$0.086 & 0.914 $\pm$0.015 & 0.785$\pm$0.012 & 0.809$\pm$0.016 &  0.968$\pm$0.004 &  0.939 $\pm$0.027 \\ \midrule
\begin{tabular}[c]{@{}l@{}}\textbf{FrontierX-} \\ \textbf{W/GPT-3.0}\end{tabular} & 0.885$\pm$0.084 &  0.536$\pm$0.173 & 13.867 $\pm$0.137 & 0.896 $\pm$0.034 & 0.774$\pm$0.028 & 0.823$\pm$0.034 &  1.025$\pm$0.007 &  0.914 $\pm$0.041\\ \midrule
\begin{tabular}[c]{@{}l@{}}\textbf{FrontierX-} \\ \textbf{W/Google Bard}\end{tabular} & 0.749$\pm$0.109 & 0.426$\pm$0.138 & 16.729 $\pm$0.145 & 0.863 $\pm$0.018 & 0.692$\pm$0.037 & 0.727$\pm$0.045 & 2.345$\pm$0.012 & 0.855 $\pm$0.017\\ \bottomrule
\bottomrule
\end{tabular}
\end{adjustbox}
\label{text_based_moldesign_results2}
\vspace{-3mm}
\end{table}

\vspace{-3mm}
\subsection{Ablation Studies}
\vspace{-3mm}
Our proposed framework operates through a series of interconnected stages via a progressively structured multi-step pipeline. Beginning with step (a), we create knowledge-augmented prompts using task-specific instructions and demonstrations, prompting large language models (LLMs) to (\romannum{1}) generate top-ranked (top-$R$) SMILES strings predictions along with (\romannum{2}) explanatory justifications for their predictions. In step (b), these generated explanations are used to (\romannum{1}) fine-tune a smaller pre-trained language model ($\textrm{LM}_{\textrm{exp}}$) for domain customization to obtain contextualized token embeddings and utilize (\romannum{2}) a weighted sum-pooling attention mechanism to compute text-level embeddings denoted as $y_{\text{exp}}$ from the token embeddings for task-specific adaptation. Moving to step (c), the top-$R$ predictions from the LLMs are transformed to compute prediction embeddings $y_{\text{pred}}$. Concurrently, in step (d), we fine-tune another small-scale language model ($\textrm{LM}_{\textrm{org}}$) on the original textual descriptions of molecules to compute context-aware token embeddings, and then compute the original text-level embeddings $y_{\text{org}}$ through a weighted attention mechanism. In step (e), our proposed framework obtains a cross-modal embeddings, $y_{\text{cross}}$, through a hierarchical multi-head attention mechanism that integrates the original text-level embeddings $y_{\text{org}}$, explanatory text-level embeddings $y_{\text{exp}}$, and prediction embeddings $y_{\text{pred}}$. We conduct empirical research to understand the significance and contribution of each distinct method within the proposed framework, evaluating its learned embeddings to achieve optimal results. We perform ablation studies to assess the impact of disabling individual methods on the overall performance of our framework. To determine the contribution of each method to the framework's performance, we create various ablated variants by disabling individual methods and evaluate them using benchmark datasets for \texttt{text2mol} tasks. We choose the proposed $\texttt{FrontierX: LLM-MG}$ framework as the reference baseline for the ablation studies. Our robust strategy not only validates the efficacy of the diverse methods but also substantiates the rationale, providing a strong basis for their design choices and justifying their inclusion within the framework. The ablated variants without the explanatory text-level embeddings, prediction embeddings, and original text-level embeddings are referred to as ``w/o $y_{\text{exp}}$", ``w/o $y_{\text{pred}}$", and ``w/o $y_{\text{org}}$", respectively. The ablation study findings are summarized in Table \ref{ablation1}. All the ablation study experiments were conducted with the \textbf{FrontierX: LLM-MG} framework using the GPT-4 backbone and Scaffold sampling technique with $K=16$, by disabling certain methods as discussed earlier.

\vspace{-2mm}
\begin{table}[!ht]
\centering
\caption{The table shows the experimental findings on the ablation study. The experiments are conducted using the Scaffold sampling technique with $K$=16 and GPT-4 backbone.}
\vspace{-2mm}
\begin{adjustbox}{width=1.25\columnwidth,center}
\hspace{-5mm}\begin{tabular}{@{}|c|cccccccc|@{}}
\toprule
\toprule
\textbf{Method} & \textbf{BLEU ($\uparrow$)}                       & \textbf{Exact ($\uparrow$)}                      & \textbf{Levenshtein ($\downarrow$)}                 & \textbf{Validity ($\uparrow$)}                   & \textbf{MACCS FTS ($\uparrow$)}                  & \textbf{RDK FTS ($\uparrow$)}                    & \textbf{Morgan FTS ($\uparrow$)}                 & \textbf{FCD ($\downarrow$)}                        \\ \midrule
\begin{tabular}[c]{@{}l@{}} 
Ground-Truth \\ \end{tabular}            & 1.0          & 1.0          & 0.0          & 1.0          & 1.0          & 1.0          & 1.0          & 0         \\ \midrule
\begin{tabular}[c]{@{}l@{}}\textbf{FrontierX} \\  \end{tabular}  & \textbf{0.937$\pm$0.023} & \textbf{0.641 $\pm$0.089}  & \textbf{6.946 $\pm$0.045} & \textbf{0.975 $\pm$0.013} &  \textbf{0.947 $\pm$0.029} & \textbf{0.838$\pm$0.006} & \textbf{0.845$\pm$0.012} & \textbf{0.796$\pm$0.003} \\ \midrule
\begin{tabular}[c]{@{}l@{}}\textbf{FrontierX} \\ --- w/o $y_{\text{exp}}$\end{tabular} & 0.783$\pm$0.032 & 0.468 $\pm$0.054  & 23.526 $\pm$0.013 & 0.847 $\pm$0.025 &  0.784 $\pm$0.061 & 0.682$\pm$0.013 & 0.723$\pm$0.037 & 3.712$\pm$0.014  \\ \midrule
\begin{tabular}[c]{@{}l@{}}\textbf{FrontierX} \\ --- w/o $y_{\text{org}}$\end{tabular} & 0.741$\pm$0.032 & 0.455 $\pm$0.089  & 27.817 $\pm$0.067 & 0.819 $\pm$0.034 &  0.751 $\pm$0.037 & 0.646$\pm$0.024 & 0.695$\pm$0.016 & 4.685$\pm$0.024  \\ \midrule
\begin{tabular}[c]{@{}l@{}}\textbf{FrontierX} \\ --- w/o $y_{\text{pred}}$\end{tabular} & 0.828$\pm$0.044 & 0.543 $\pm$0.063  & 13.946 $\pm$0.045 & 0.911 $\pm$0.048 &  0.867 $\pm$0.058 & 0.749$\pm$0.019 & 0.763$\pm$0.027 & 2.736$\pm$0.033  \\  \bottomrule
\bottomrule
\end{tabular}
\end{adjustbox}
\label{ablation1}
\vspace{-2mm}
\end{table}

\vspace{-1mm}
On the ChEBI-20 dataset\cite{edwards2021text2mol}, we observe differing impacts on framework performance when certain methods are omitted. The ``w/o $y_{\text{exp}}$" variant shows a substantial decline in performance relative to the baseline, as evidenced by a significant drop of $17.21\%$ in MACCS FTS, $16.43\%$ in BLEU, and $13.12\%$ in Validity. Similarly, the ``w/o $y_{\text{org}}$" variant performs much worse than the baseline, with a remarkable drop of $20.69\%$ in MACCS FTS, $20.91\%$ in BLEU, and $16.00\%$ in Validity.  In contrast, the ``w/o $y_{\text{pred}}$" variant exhibits a marginally inferior performance compared to the baseline, with a modest drop of $8.44\%$ in MACCS FTS, $11.63\%$ in BLEU, and $6.56\%$ in Validity. The significant drop in performance metrics for the ablated variants, when compared to the baseline, highlights the considerable impact of the mechanisms inherent in the methods omitted from the baseline and leads to degraded performance. Our experiments corroborate our hypothesis of joint optimization to obtain a cross-modal embeddings, $y_{\text{cross}}$, through a hierarchical multi-head attention mechanism that integrates the original text-level embeddings $y_{\text{org}}$, explanatory text-level embeddings $y_{\text{exp}}$, and prediction embeddings $y_{\text{pred}}$, achieving state-of-the-art (SOTA) performance on the \texttt{text2mol} task

\vspace{-5mm}
\section{Conclusion}
\vspace{-4mm}
In this study, we pioneered the \texttt{text2mol} approach, inaugurating a transformative paradigm where chemistry meets language models, expediting scientific advancements. Through the creation of \texttt{FrontierX: LLM-MG}, we demonstrated the efficacy of using large language models for seamless and efficient translation between textual descriptions and chemical SMILES representations. Acknowledging the limitations of current methods, our research highlights a promising horizon in molecule design, potentially ushering in an era of accelerated innovation and interdisciplinary collaboration. Our study illustrates the transformative impact of integrating molecular design with language models, offering an innovative approach to molecule generation that can catalyze groundbreaking developments in science and technology.
\newpage

\section{Technical Appendix} 
\vspace{-3mm}
\subsection{Study of Knowledge-Augmented Prompting} 
\vspace{-2mm}
In our study, we employ knowledge-augmented prompting with LLMs for zero-shot text-to-molecule(\texttt{text2mol}) translation task by leveraging the pre-existing knowledge embedded within the language model parameters. LLMs are capable of generating chemical SMILES representations from textual descriptions through entity recognition, grammar understanding, symbol mapping, and structure validation, which marks significant progress in molecule generation via language models. This knowledge-augmentation prompting technique allows LLMs to adapt to new, unseen molecule textual descriptions using a few task-specific demonstrations, thereby eliminating the need for fine-tuning with labeled data for task-specific adaptation. The approach involves creating knowledge-augmented prompts that combine task-specific instructions with demonstrations (input-output pairs) sampled from training data relevant to the target molecule textual descriptions determined using off-the-shelf semantic similarity techniques. In this context, each pair consists of a textual description of a molecule (input) and its corresponding SMILES representation (output), where the task-specific instruction is to convert the target molecule textual descriptions into the standardized chemical SMILES notation. This approach aligns the LLM's capabilities with the \texttt{text2mol} task by crafting knowledge-augmented prompts that blend specific instructions with relevant demonstrations, selected based on semantic similarity. This strategic alignment facilitates accurate chemical SMILES strings generation without necessitating language model parameter updates. We have employed two sampling strategies --- random and scaffold --- to evaluate the impact of both the quality and quantity of demonstrations in the knowledge-augmented prompt, which is utilized for querying LLMs during text-based de novo molecule generation. The scaffold strategy utilizes a semantic similarity method to sample the top-$K$ relevant text-molecule pairs, using OpenAI's \texttt{text-embedding-ada-002 technique}\footnote{https://platform.openai.com/docs/guides/embeddings}. The random technique involves the arbitrary selection of $K$ text-molecule pairs without any prior knowledge in a non-deterministic manner. The study compares the effectiveness of both strategies in enhancing the language-conditioned molecule generation task using off-the-shelf pre-trained LLMs, including Google Bard and other GPT model family variants. We conducted experiments to compare and contrast the performance of the ``Random'' and ``Scaffold'' sampling strategies, and to identify the optimal number of demonstrations.

\vspace{-2mm}
\begin{table}[!ht]
\centering
\caption{The table shows the results of the experimental study examining the impact of both quantity and quality of demonstrations on knowledge-augmented prompting strategies in the \texttt{text2mol} task.}
\vspace{-2mm}
\begin{adjustbox}{width=1.25\columnwidth,center}
\hspace{-5mm}\begin{tabular}{@{}|c|cccccccc|@{}}
\toprule
\toprule
\textbf{Method} & \textbf{BLEU ($\uparrow$)}                       & \textbf{Exact ($\uparrow$)}                      & \textbf{Levenshtein ($\downarrow$)}                 & \textbf{Validity ($\uparrow$)}                   & \textbf{MACCS FTS ($\uparrow$)}                  & \textbf{RDK FTS ($\uparrow$)}                    & \textbf{Morgan FTS ($\uparrow$)}                 & \textbf{FCD ($\downarrow$)}                        \\ \midrule
\begin{tabular}[c]{@{}l@{}} 
Ground-Truth \\ \end{tabular}            & 1.0          & 1.0          & 0.0          & 1.0          & 1.0          & 1.0          & 1.0          & 0         \\ \midrule
\begin{tabular}[c]{@{}l@{}}\textbf{FrontierX: LLM-MG } \\ - (Baseline - Scaffold, $k$=16, W/GPT-4) \end{tabular}  & \textbf{0.937$\pm$0.023} & \textbf{0.641 $\pm$0.089}  & \textbf{6.946 $\pm$0.045} & \textbf{0.975 $\pm$0.013} &  \textbf{0.947 $\pm$0.029} & \textbf{0.838$\pm$0.006} & \textbf{0.845$\pm$0.012} & \textbf{0.796$\pm$0.003} \\ \midrule
\begin{tabular}[c]{@{}l@{}}\textbf{FrontierX: LLM-MG } \\ - (Scaffold, $k$=12, W/GPT-3.5-turbo) \end{tabular} & 0.837$\pm$0.067 & 0.527$\pm$0.121 & 14.117 $\pm$0.072 & 0.885 $\pm$0.067 & 0.868 $\pm$0.052 & 0.773$\pm$0.026 & 0.763$\pm$0.037 &  2.099$\pm$0.045 \\ \midrule
\begin{tabular}[c]{@{}l@{}}\textbf{FrontierX: LLM-MG } \\ - (Scaffold, $k$=12, W/GPT-3.0-text-davinci-003)\end{tabular} & 0.843$\pm$0.032 &  0.487$\pm$0.135 & 16.789 $\pm$0.096 & 0.867 $\pm$0.059 & 0.856 $\pm$0.102 & 0.714$\pm$0.042 & 0.776$\pm$0.064 &  3.242$\pm$0.014\\ \midrule
\begin{tabular}[c]{@{}l@{}}\textbf{FrontierX: LLM-MG} \\ - (Scaffold, $k$=12, W/Google Bard)\end{tabular} & 0.703$\pm$0.128 & 0.466$\pm$0.104 & 19.211 $\pm$0.145 & 0.803 $\pm$0.074 & 0.819 $\pm$0.082 & 0.614$\pm$0.075 & 0.687$\pm$0.067 & 3.105$\pm$0.054\\ \midrule
\begin{tabular}[c]{@{}l@{}}\textbf{FrontierX: LLM-MG} \\ - (random, $k$=12,  W/GPT-3.5-turbo)\end{tabular} & 0.735$\pm$0.033 & 0.417$\pm$0.102 & 21.324 $\pm$0.089 & 0.741 $\pm$0.097 & 0.767 $\pm$0.085 & 0.692$\pm$0.063 & 0.667$\pm$0.059 &  8.125$\pm$0.075 \\ \midrule
\begin{tabular}[c]{@{}l@{}}\textbf{FrontierX: LLM-MG} \\ - (random, $k$=12, W/GPT-3.0-text-davinci-003)\end{tabular} & 0.716$\pm$0.045 &  0.327$\pm$0.087 & 23.196 $\pm$0.054 & 0.753 $\pm$0.109 & 0.766 $\pm$0.113 & 0.609$\pm$0.012 & 0.684$\pm$0.019 &  9.309$\pm$0.014\\ \midrule
\begin{tabular}[c]{@{}l@{}}\textbf{FrontierX: LLM-MG} \\ - (random, $k$=12, W/Google Bard)\end{tabular} & 0.623$\pm$0.113 & 0.403$\pm$0.064 & 15.103 $\pm$0.165 & 0.717 $\pm$0.126 & 0.689 $\pm$0.037 & 0.504$\pm$0.085 & 0.597$\pm$0.073 & 15.135$\pm$0.057\\ \midrule
\begin{tabular}[c]{@{}l@{}}\textbf{FrontierX: LLM-MG} \\ - (Scaffold, $k$=4, W/GPT-3.5-turbo)\end{tabular} & 0.605$\pm$0.077 & 0.346$\pm$0.061 & 26.106 $\pm$0.081 & 0.651 $\pm$0.079 & 0.697 $\pm$0.035 & 0.612$\pm$0.085 & 0.577$\pm$0.031 &  17.321$\pm$0.056 \\ \midrule
\begin{tabular}[c]{@{}l@{}}\textbf{FrontierX: LLM-MG} \\ - (Scaffold, $k$=4, W/GPT-3.0-text-davinci-003)\end{tabular} & 0.623$\pm$0.042 &  0.259$\pm$0.038 & 26.096 $\pm$0.037 & 0.673 $\pm$0.068 & 0.686 $\pm$0.093 & 0.517$\pm$0.027 & 0.603$\pm$0.061 &  19.219$\pm$0.038\\ \midrule
\begin{tabular}[c]{@{}l@{}}\textbf{FrontierX: LLM-MG} \\ - (Scaffold, $k$=4, W/Google Bard)\end{tabular}  & 0.514$\pm$0.049& 0.317$\pm$0.047 & 22.033 $\pm$0.187 & 0.636 $\pm$0.137 & 0.587 $\pm$0.056 & 0.446$\pm$0.032 & 0.508$\pm$0.049 & 24.101$\pm$0.095 \\ \bottomrule
\bottomrule
\end{tabular}
\end{adjustbox}
\label{knowledgeaug}
\vspace{-2mm}
\end{table}

\vspace{-3mm}
\paragraph{Results:} Table \ref{knowledgeaug} presents the results of the experimental study that examined the effects of both the quantity and quality of demonstrations on the performance of knowledge-augmented prompting strategies in the \texttt{text2mol} task. Our study compared the performance of various GPT models with that of Google Bard on the ChEBI-20 dataset\cite{edwards2021text2mol}. The results indicated that the GPT models consistently outperformed Google Bard across all evaluation metrics when provided with the same number of task-specific demonstrations in the augmented prompt. Notably, GPT-4 demonstrated the highest performance among the tested models, generating a greater number of valid chemical SMILES representations. Furthermore, the study indicates that enhancing the knowledge-augmented prompt with more task-specific demonstrations directly improves the predictive accuracy of the language models. This highlights a positive correlation between the number of task-specific demonstrations and the performance of LLMs on the \texttt{text2mol} task. The study found that scaffold sampling consistently outperforms random sampling on the \texttt{text2mol} task when using any off-the-shelf LLMs. One possible reason for this superior performance is the strong textual similarities between the text-molecule pairs sampled using the scaffold technique and the target molecule descriptions. Therefore, using scaffold sampling instead of random sampling may lead GPT models to generate more accurate chemical SMILES representations. LLMs continue to face challenges in precisely interpreting molecular representations in chemical SMILES notations, resulting in poor performance on \texttt{text2mol} tasks. SMILES representations can possess multiple valid forms and implicit hydrogen atoms, causing ambiguity and presenting difficulties for LLMs. Improved LLMs capable of handling molecular structures and seamlessly integrating with tools like RDKit are necessary

\vspace{-4mm}
\subsection{Impact of Hierarchical Multi-Head Attention(HMHA) Mechanism}
\vspace{-3mm} 
In our work, we compute the cross-modal embedding, denoted as $y_{\text{cross}},$ through a hierarchical multi-head attention (HMHA) mechanism that integrates the original text-level embedding ($y_{\text{org}}$), explanatory text-level embedding ($y_{\text{exp}}$), and prediction embedding ($y_{\text{pred}}$). To determine the impact of the HMHA mechanism on the framework performance, we conducted ablation study. We refer to the ablated variant without the HMHA mechanism as ``w/o $\text{HMHA}$''. We substitute the HMHA mechanism with dual-stage linear operators in the ablated variant to compute cross-modal embeddings. The findings of the ablation study are summarized in Table \ref{ablation3}. We conducted the experiment using the \texttt{FrontierX: LLM-MG} framework with GPT-4 backbone, where we replaced the $\text{HMHA}$ mechanism with linear operators, as discussed earlier. The experimental results support the inclusion of the hierarchical multi-head attention mechanism (HMHA) to generate cross-modal embeddings, aiding in the generation of more valid chemical SMILES representations in the \texttt{text2mol} task.

\vspace{-2mm}
\begin{table}[!ht]
\centering
\caption{The table shows the experimental findings of the study on the impact of the HMHA mechanism on the \texttt{text2mol} task. The experiments were conducted using the \texttt{FrontierX: LLM-MG} framework with a GPT-4 backbone. We utilized the Scaffold sampling technique with $K=16$ for constructing augmented prompts.}
\vspace{-2mm}
\begin{adjustbox}{width=1.25\columnwidth,center}
\hspace{-5mm}\begin{tabular}{@{}|c|cccccccc|@{}}
\toprule
\toprule
\textbf{Method} & \textbf{BLEU ($\uparrow$)}                       & \textbf{Exact ($\uparrow$)}                      & \textbf{Levenshtein ($\downarrow$)}                 & \textbf{Validity ($\uparrow$)}                   & \textbf{MACCS FTS ($\uparrow$)}                  & \textbf{RDK FTS ($\uparrow$)}                    & \textbf{Morgan FTS ($\uparrow$)}                 & \textbf{FCD ($\downarrow$)}                        \\ \midrule
\begin{tabular}[c]{@{}l@{}} 
Ground-Truth \\ \end{tabular}            & 1.0          & 1.0          & 0.0          & 1.0          & 1.0          & 1.0          & 1.0          & 0         \\ \midrule
\begin{tabular}[c]{@{}l@{}}\textbf{FrontierX} \\ - (W/GPT-4) \end{tabular} & \textbf{0.937$\pm$0.023} & \textbf{0.641 $\pm$0.089}  & \textbf{6.946 $\pm$0.045} & \textbf{0.975 $\pm$0.013} &  \textbf{0.947 $\pm$0.029} & \textbf{0.838$\pm$0.006} & \textbf{0.845$\pm$0.012} & \textbf{0.796$\pm$0.003} \\ \midrule

\begin{tabular}[c]{@{}l@{}}\textbf{FrontierX} \\ - (W/GPT-4) -  w/o $\textbf{HMHA}$\end{tabular} & \text{0.773$\pm$0.033} & \text{0.516 $\pm$0.069}  & \text{22.657 $\pm$0.082} & \text{0.787 $\pm$0.037} &  \text{0.756 $\pm$0.074} & \text{0.691$\pm$0.017} & \text{0.694$\pm$0.027} & \text{5.598$\pm$0.065} \\ \bottomrule
\bottomrule
\end{tabular}
\end{adjustbox}
\label{ablation3}
\vspace{-2mm}
\end{table}

\vspace{-2mm}
\begin{table}[!ht]
\centering
\caption{The table shows the experimental results of the hyperparameter tuning experiments.}
\vspace{-2mm}
\begin{adjustbox}{width=1.2\columnwidth,center}
\hspace{-5mm}\begin{tabular}{@{}|c|cccccccc|@{}}
\toprule
\toprule
\textbf{Method} & \textbf{BLEU ($\uparrow$)}                       & \textbf{Exact ($\uparrow$)}                      & \textbf{Levenshtein ($\downarrow$)}                 & \textbf{Validity ($\uparrow$)}                   & \textbf{MACCS FTS ($\uparrow$)}                  & \textbf{RDK FTS ($\uparrow$)}                    & \textbf{Morgan FTS ($\uparrow$)}                 & \textbf{FCD ($\downarrow$)}                        \\ \midrule
\begin{tabular}[c]{@{}l@{}} 
Ground-Truth \\ \end{tabular}            & 1.0          & 1.0          & 0.0          & 1.0          & 1.0          & 1.0          & 1.0          & 0         \\ \midrule
\begin{tabular}[c]{@{}l@{}}\textbf{FrontierX: LLM-MG } \\ - ($b$=32, $d$=128) \end{tabular}  & \textbf{0.937$\pm$0.023} & \textbf{0.641 $\pm$0.089}  & \textbf{6.946 $\pm$0.045} & \textbf{0.975 $\pm$0.013} &  \textbf{0.947 $\pm$0.029} & \textbf{0.838$\pm$0.006} & \textbf{0.845$\pm$0.012} & \textbf{0.796$\pm$0.003} \\ \midrule
\begin{tabular}[c]{@{}l@{}}\textbf{FrontierX: LLM-MG } \\ - ($b$=48, $d$=196)\end{tabular} & \text{0.887$\pm$0.017} & \text{0.604 $\pm$0.094}  & \text{8.581 $\pm$0.084} & \text{0.942 $\pm$0.019} &  \text{0.895 $\pm$0.094} & \text{0.797$\pm$0.012} & \text{0.817$\pm$0.018} & \text{0.827$\pm$0.015} \\ \midrule
\begin{tabular}[c]{@{}l@{}}\text{FrontierX: LLM-MG } \\ - ($b$=48, $d$=128) \end{tabular} & \text{0.893$\pm$0.023} & \text{0.627 $\pm$0.081}  & \text{8.024 $\pm$0.053} & \text{0.955 $\pm$0.022} &  \text{0.915 $\pm$0.044} & \text{0.807$\pm$0.067} & \text{0.829$\pm$0.025} & \text{0.814$\pm$0.033} \\ \midrule
\begin{tabular}[c]{@{}l@{}}\textbf{FrontierX: LLM-MG} \\ - ($b$=64, $d$=256) \end{tabular} & \text{0.903$\pm$0.011} & \text{0.633 $\pm$0.072}  & \text{7.813 $\pm$0.047} & \text{0.915 $\pm$0.032} &  \text{0.929 $\pm$0.036} & \text{0.817$\pm$0.069} & \text{0.833$\pm$0.037} & \text{0.821$\pm$0.037} \\ \bottomrule
\bottomrule
\end{tabular}
\end{adjustbox}
\label{hypertune}
\vspace{-3mm}
\end{table}

\vspace{-3mm}
\subsection{Hyperparameter Tuning}
\vspace{-3mm}
To enhance the performance of our \texttt{FrontierX: LLM-MG} framework, we embarked on meticulous hyperparameter tuning through detailed experimentation and analysis. We chose random search as a strategy to adeptly navigate the hyperparameter space, pinpointing the optimal framework configuration on the benchmark dataset, in lieu of more computationally intensive approaches such as grid search or Bayesian optimization. This strategy enabled us to obtain optimal results on the validation subset of the benchmark dataset, as evidenced by several evaluation metrics. We conducted hyperparameter optimization on the \texttt{FrontierX: LLM-MG-W/GPT-4} variant of our framework. We utilized the Scaffold sampling technique with K = 16 for constructing augmented prompts.The primary key hyperparameters within this framework include batch size ($b \in \{32, 48, 64 \}$) and the embedding dimension ($d \in \{64, 128, 196, 256\}$). Table \ref{hypertune} presents the results of hyperparameter tuning on the representative benchmark dataset. We report the results for the near-optimal combinations of the hyperparameters.

\vspace{-3mm}
\subsection{Molecule captioning}
\vspace{-2mm}
Molecule captioning is a crucial task in the field of computational chemistry, serving as a bridge between complex chemical data and human comprehension. It involves generating detailed and correct textual descriptions that accurately describe a chemical SMILES representation in the \texttt{mol2text} task. This stands in contrast to the \texttt{text2mol} task, which entails generating chemical SMILES representations from detailed and factual textual descriptions. Meanwhile, the \texttt{mol2text} task helps to translate complex chemical structures into understandable language, enhancing our understanding of molecules with potential applications spanning multiple fields, including drug discovery, materials science, and chemical synthesis. To evaluate the quality of the generated text in the \texttt{mol2text} task, we employ traditional metrics commonly used in natural language processing and machine translation, including BLEU, ROUGE, and METEOR, as described below:

\vspace{-2mm}
\begin{itemize}
    \item \textbf{BLEU-2} and \textbf{BLEU-4} are part of the BiLingual Evaluation Understudy (BLEU) metric family. BLEU is typically computed for different n-gram levels, where `n' represents the number of contiguous words or tokens considered. BLEU-2 evaluates the accuracy of two-word phrases (bigrams) in generated text, while BLEU-4 extends this analysis to four-word sequences (4-grams). These metrics offer insights into the alignment between machine-generated and human reference texts. The BLEU metric variants help quantify the performance of language generation-based NLP models.    
   \item \textbf{ROUGE-1} and \textbf{ROUGE-2} are part of the Recall-Oriented Understudy for Gisting Evaluation (ROUGE) metric family. ROUGE-1, also known as ROUGE unigram or ROUGE-N1, evaluates the overlap between generated and reference text at the unigram (single-word) level, assessing specific word choices. In contrast, ROUGE-2 (also known as ROUGE bigram or ROUGE-N2) extends this to evaluate consecutive word pairs (bigrams), offering comprehensive insights into content matching. These metrics measure alignment with reference texts at both the word and bigram levels, providing precision and recall evaluations of textual elements.    
    \item \textbf{ROUGE-L}, or Recall-Oriented Understudy for Gisting Evaluation - Longest Common Subsequence, measures the quality of machine-generated text by considering the longest common subsequence between the generated text and a reference text. This subsequence represents a sequence of words that appear in the same order in both the generated and reference texts, allowing flexibility in word order. ROUGE-L assesses content overlap and structural similarity, capturing the core content and organization of generated text concerning the reference text, even when there are variations in wording or word order.
    \item  \textbf{METEOR} (Metric for the Evaluation of Translation with Explicit Ordering) considers precision, recall, stemming, synonymy, and word order, offering a well-rounded evaluation of text quality by analyzing matching words and their order, providing detailed assessments for translation and captioning models. 
\end{itemize}

\vspace{-4mm}
\subsubsection{FrontierX: LLM-MG - \texttt{mol2text} task}
\begin{figure}[ht!]
\centering
\resizebox{1.0\linewidth}{!}{ 
\hspace*{-3mm}\includegraphics[keepaspectratio,height=5.0cm,trim=0.0cm 3.0cm 0cm 2.0cm,clip]{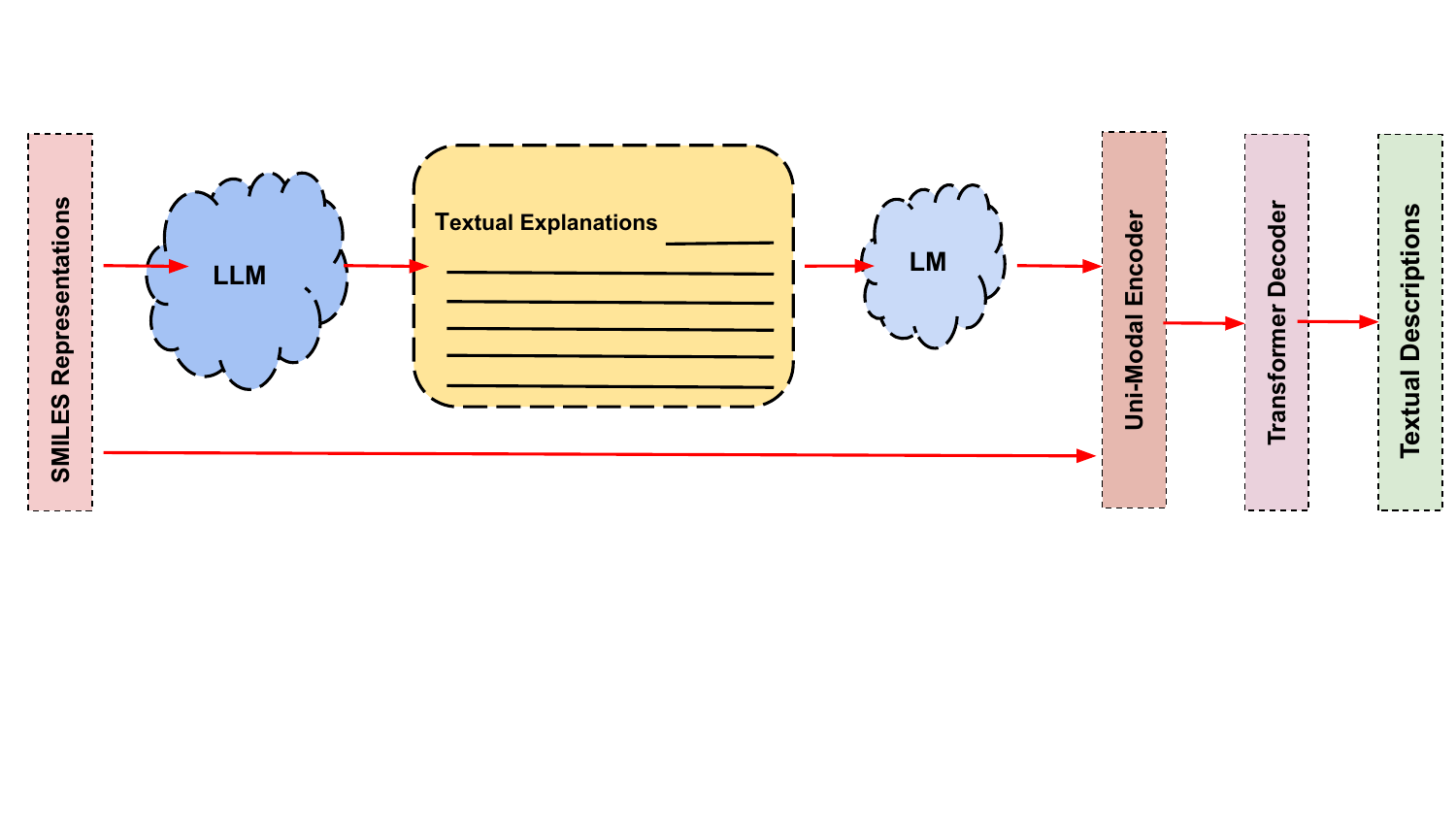} 
}
\vspace{-18mm}
\caption{Overview of \textbf{FrontierX: LLM-MG} framework for \texttt{mol2text} task.}
\label{fig:figure2}
\end{figure}

\vspace{-2mm}
We have modified the FrontierX: LLM-MG pipeline for the \texttt{text2mol} task to adapt it for the \texttt{mol2text} task. The workflow of the proposed approach is illustrated in Figure \ref{fig:figure2}. Given a chemical SMILES representation, it can generate the technical descriptions of the molecule. We construct a knowledge-infused prompt using task-specific instructions and a few demonstrations (input-output mappings) for the downstream \texttt{mol2text} task. The task-specific instructions involve translating chemical SMILES representations into their corresponding technical descriptions. The primary objective of the prompt engineering method is to enhance the context-awareness of language models and improve their ability to provide relevant and accurate responses. This enhancement is achieved through learning from demonstrations, rather than relying on conventional supervised learning methods of fine-tuning with labeled data for the \texttt{mol2text} task. The knowledge-infused prompts guide the language models to generate technical descriptions based on the provided instructions. Next, we fine-tune small-scale, pre-trained language models (LMs) using the generated explanations, which facilitates domain customization and yields context-aware token embeddings. To create a text-level embedding that encapsulates the generated technical descriptions, we utilize a weighted sum-pooling attention mechanism on the contextualized embeddings. Additionally, the unimodal encoder, which is implemented with a multi-head attention mechanism, integrates the mono-domain chemical SMILES representations with explanatory text-level embeddings to compute unimodal embeddings. Finally, the transformer decoder generates the technical descriptions that correspond to the input chemical SMILES representations\cite{vaswani2017attention}. Instead of repurposing large language models (LLMs) by either retraining them from scratch or fine-tuning them with labeled data for domain customization, we employ LMaaS\cite{sun2022black} to engage with LLMs through text-based API interactions.

\vspace{-3mm}

\vspace{-2mm}
\begin{table}[!ht]
\centering
\caption{The table shows the experimental findings of the framework performance on the \texttt{mol2text} task in comparison with the baselines. }
\vspace{-1mm}
\scalebox{0.8}{
\hspace*{-5mm}\begin{tabular}{l|c|c|c|c|c|c}
\toprule
Method                                                                  & BLEU-2 ($\uparrow$)                     & BLEU-4 ($\uparrow$)                     & ROUGE-1 ($\uparrow$)                    & ROUGE-2 ($\uparrow$)                    & ROUGE-L ($\uparrow$)                   & METEOR ($\uparrow$)                     \\ \midrule
\begin{tabular}[c]{@{}l@{}} GPT-4 \\ (zero-shot)\end{tabular}            & 0.062$\pm$0.001          & 0.013$\pm$0.001          & 0.192$\pm$0.002          & 0.040$\pm$0.002          & 0.125$\pm$0.002          & 0.209$\pm$0.002          \\ \midrule
\begin{tabular}[c]{@{}l@{}}GPT-4 \\ (Scaffold, $k$=10)\end{tabular}       & 0.464$\pm$0.008          & 0.365$\pm$0.008          & 0.545$\pm$0.003 & 0.362$\pm$0.003 & 0.459$\pm$0.007          & 0.519$\pm$0.005 \\ \midrule
\begin{tabular}[c]{@{}l@{}}GPT-4 \\ (Scaffold, $k$=5)\end{tabular}        & 0.456$\pm$0.003          & 0.357$\pm$0.004          & 0.540$\pm$0.005          & 0.355$\pm$0.007          & 0.455$\pm$0.005          & 0.505$\pm$0.005          \\ \midrule
\begin{tabular}[c]{@{}l@{}}GPT-4 \\ (Random, $k$=10)\end{tabular}         & 0.260$\pm$0.007          & 0.140$\pm$0.007          & 0.393$\pm$0.004          & 0.180$\pm$0.006          & 0.309$\pm$0.004          & 0.320$\pm$0.007          \\ \midrule
\begin{tabular}[c]{@{}l@{}}GPT-3.5 \\ (Scaffold, $k$=10)\end{tabular}     & 0.468$\pm$0.010          & 0.368$\pm$0.010          & 0.534$\pm$0.005          & 0.355$\pm$0.007          & 0.457$\pm$0.006          & 0.497$\pm$0.005          \\ \midrule
\begin{tabular}[c]{@{}l@{}}Davinci-003 \\ (Scaffold, $k$=10)\end{tabular} & 0.488$\pm$0.011 & 0.391$\pm$0.012 & 0.532$\pm$0.008          & 0.359$\pm$0.010          & 0.465$\pm$0.008 & 0.478$\pm$0.011          \\  \midrule
\begin{tabular}[c]{@{}l@{}}\textbf{FrontierX: LLM-MG } \\ (Scaffold, $k$=16, W/GPT-4) \end{tabular} & \textbf{0.743 $\pm$0.081} & \textbf{0.656 $\pm$0.097} & \textbf{0.818 $\pm$0.034}       &     \textbf{0.727 $\pm$0.013}     & \textbf{0.783 $\pm$0.047} &      \textbf{0.812 $\pm$0.051}     \\  \bottomrule
\end{tabular}
}
\label{captioning_results1}
\vspace{-2mm}
\end{table}

\vspace{-2mm}
\begin{table}[!ht]
\centering
\caption{The table presents the framework performance and the baselines on the experimental study on the \texttt{mol2text} task.}
\vspace{-1mm}
\scalebox{0.8}{
\hspace*{-5mm}\begin{tabular}{l|c|c|c|c|c|c}
\toprule
Method                                                                  & BLEU-2 ($\uparrow$)                     & BLEU-4 ($\uparrow$)                     & ROUGE-1 ($\uparrow$)                    & ROUGE-2 ($\uparrow$)                    & ROUGE-L ($\uparrow$)                   & METEOR ($\uparrow$)                     \\ \midrule
RNN & 0.251 & 0.176 & 0.450 & 0.278 & 0.394 & 0.363  \\  \midrule
Transformer & 0.061 & 0.027 & 0.204 & 0.087 & 0.186 & 0.114  \\   \midrule
T5-Small & 0.501 & 0.415 & 0.602 & 0.446 & 0.545 & 0.532   \\  \midrule
MolT5-Small & 0.519 & 0.436 & 0.620 & 0.469 & 0.563 & 0.551   \\ \midrule
T5-Base & 0.511 & 0.423 & 0.607 & 0.451 & 0.550 & 0.539   \\  \midrule
MolT5-Base & 0.540 & 0.457 & 0.634 & 0.485 & 0.578 & 0.569   \\ \midrule
T5-Large & 0.558 & 0.467 & 0.630 & 0.478 & 0.569 & 0.586   \\  \midrule
MolT5-Large & 0.594 & 0.508 & 0.654 & 0.510 & 0.594 & 0.614  \\  \midrule
\begin{tabular}[c]{@{}l@{}}\textbf{FrontierX: LLM-MG } \\ (Scaffold, $k$=16, W/GPT-4) \end{tabular} & \textbf{0.743 $\pm$0.081} & \textbf{0.656 $\pm$0.097} & \textbf{0.818 $\pm$0.034}       &     \textbf{0.727 $\pm$0.013}     & \textbf{0.783 $\pm$0.047} &      \textbf{0.812 $\pm$0.051}     \\  \bottomrule
\end{tabular}
}
\label{captioning_results2}
\vspace{-2mm}
\end{table}
 
\vspace{-3mm}
\subsubsection{LLM Prompting}
\vspace{-1mm}
Knowledge-infused LLM prompting, a method of prompt engineering, involves crafting effective prompts or input queries to elicit desired responses from language models. This technique enhances language models by combining their natural language understanding and generation capabilities with access to external factual information(demonstrations), making them more versatile for task-specific applications. Consequently, it enables the LLM to generate responses enriched with accurate, contextually relevant information. Knowledge-infused prompting enables LLMs to adapt to new tasks without the need for explicit, gradient-based fine-tuning with gold-standard annotated data for task-specific adaptation in downstream tasks \cite{brown2020language}). This approach allows LLMs to acquire knowledge through analogies, relying on a limited set of input-output mappings (demonstrations) tailored to the specific downstream task. Knowledge-infused prompting harnesses the implicit knowledge embedded in pretrained LLM parameters to facilitate adaptation to new tasks via task-specific demonstrations, all without necessitating parameter updates. The Knowledge-Infused prompt provides task-specific instructions and demonstrations, allowing LLMs to generate outputs conditioned on the prompt for improved generalization performance. In the case of \texttt{mol2text} tasks, we construct a knowledge-infused prompt using a few demonstrations sampled from the training data. To examine how the quality and quantity of task-specific demonstrations impact performance on \texttt{mol2text} tasks, we investigate two different sampling strategies. The quality of examples is determined by the retrieval techniques employed to select the top-$K$ demonstrations (chemical SMILES strings-text data pairs) from the training set that match the query chemical SMILES representations. We explore two distinct sampling strategies: `Random' and `Scaffold'. To study the impact of the quantity of demonstrations on the framework's performance on the \texttt{mol2text} task, we optimize the number of demonstrations ($K$) used to construct the augmented prompt for each query chemical SMILES representation. In the `Random' strategy, we randomly sample $K$ demonstrations from the training data. In contrast, the `Scaffold' strategy uses Tanimoto similarity \cite{tanimoto1958elementary} based on Morgan fingerprints \cite{morgan1965generation} with a radius of 2 to identify the top-$K$ most similar chemical SMILES representations from the training data for query chemical SMILES representations. We explore the different sampling strategies to analyze the impact of the quality of demonstrations on the \texttt{mol2text} task with a hypothesis that the `Scaffold' sampling technique outperforms the `Random' technique for the same number of demonstrations. In summary, our goal is to task LLMs with a knowledge-infused prompt that consists of a few demonstrations for the \texttt{mol2text} task, along with task-specific instructions, where the output is technical descriptions of the query chemical SMILES representation. The task-specific instruction in the augmented prompt guides LLMs to generate technical descriptions. This task showcases the LLM's capacity to generate textual descriptions via prompt conditioning, relying on its inherent knowledge, without requiring parameter updates, in contrast to supervised learning, which relies on labeled data for parameter updates. Tables \ref{captioning_results1} and \ref{captioning_results2} present the experimental findings on the ChEBI-20 benchmark dataset\cite{edwards2021text2mol}. We report the baseline results from earlier studies \cite{guo2023indeed, edwards2022translation}. The best performing model is in bold font.

\vspace{-4mm}
\subsubsection{Ablation Studies}
\vspace{-2mm}
Our proposed framework operates in a structured, multi-step pipeline. In step (a), we create knowledge-augmented prompts using task-specific instructions and demonstrations, prompting large language models (LLMs) to generate textual descriptions. In step (b), we use these generated explanations to fine-tune a smaller, pre-trained language model ($\textrm{LM}_{\textrm{exp}}$) for domain-specific customization, resulting in context-sensitive token embeddings. We employ a weighted sum-pooling attention mechanism for task-specific adaptation to compute text-level embeddings, denoted as $y_{\text{exp}}$, from the contextualized token embeddings. In parallel, in step (c), we fine-tune another small-scale language model ($\textrm{LM}_{\textrm{org}}$) on query chemical SMILES representations, computing an entire chemical SMILES string embeddings $y_{\text{org}}$. In step (d), our framework obtains a unimodal embedding, $y_{\text{uni}}$, through a multi-head attention mechanism that integrates the original text-level embeddings $y_{\text{org}}$ and descriptive text-level embeddings $y_{\text{exp}}$.  In the final step, the transformer decoder generates the textual descriptions of the query chemical SMILES string from the unimodal embedding, $y_{\text{uni}}$. Our empirical research aims to elucidate the significance and unique contributions of each method within our proposed framework, particularly in assessing the effectiveness of their learned embeddings for achieving optimal results on the \texttt{mol2text} task. Ablation studies have been conducted to investigate the impact of disabling individual methods on our framework's overall performance in the \texttt{mol2text} task. To precisely measure the impact of each method on the framework's performance, we have generated various ablated variants by disabling individual methods and assessed their performance using a benchmark dataset across multiple evaluation metrics for the \texttt{mol2text} task. We choose the $\texttt{FrontierX: LLM-MG}$ framework as the reference baseline for ablation studies in the context of the \texttt{mol2text} task. Our comprehensive approach not only confirms the effectiveness of various methods but also provides substantial support for their design choices, reinforcing their rationale and justifying their inclusion in the framework. The ablated variants without the descriptive text-level embeddings and the original text-level embeddings are denoted as `w/o $y_{\text{exp}}$' and `w/o $y_{\text{org}}$', respectively. The findings of the ablation study are summarized in Table \ref{ablation2}. All ablation experiments were conducted with the \texttt{FrontierX: LLM-MG W/GPT-4} framework using the Scaffold sampling technique with a value of $K=16$, involving the deliberate exclusion of specific methods, as previously described. On the ChEBI-20 dataset\cite{edwards2021text2mol}, the `w/o $y_{\text{exp}}$' variant exhibits a significant decline in performance relative to the baseline, evidenced by a $10.63\%$ drop in BLEU-2, a $14.04\%$ drop in ROUGE-L, and a $20.93\%$ drop in METEOR. Similarly, the `w/o $y_{\text{org}}$' variant performs much worse than the baseline, with a $20.18\%$ drop in BLEU-2, a $25.06\%$ drop in ROUGE-L, and a $29.06\%$ drop in METEOR. These substantial performance drops across all evaluation metrics when comparing the ablated variants to the baseline consistently highlighting the significant impact of the mechanisms disabled from the baseline. Our experiments validate our hypothesis on joint optimization to obtain a unimodal embedding, denoted as $y_{\text{uni}}$, through a multi-head attention mechanism that combines the original text-level embeddings $y_{\text{org}}$ and explanatory text-level embeddings $y_{\text{exp}}$. This approach results in achieving state-of-the-art (SOTA) performance on the \texttt{mol2text} task.

\vspace{-1mm}
\begin{table}[!ht]
\centering
\caption{The table shows the experimental findings on the ablation study on the \texttt{mol2text} task.}
\vspace{-2mm}
\begin{adjustbox}{width=1.075\columnwidth,center}
\hspace{-5mm}\begin{tabular}{@{}|c|cccccccc|@{}}
\toprule
\toprule
Method                                                                  & BLEU-2 ($\uparrow$)                     & BLEU-4 ($\uparrow$)                     & ROUGE-1 ($\uparrow$)                    & ROUGE-2 ($\uparrow$)                    & ROUGE-L ($\uparrow$)                   & METEOR ($\uparrow$)                     \\ \midrule
\begin{tabular}[c]{@{}l@{}}\textbf{FrontierX: LLM-MG } \\ (Scaffold, $k$=16, W/GPT-4) \end{tabular} & \textbf{0.743 $\pm$0.081} & \textbf{0.656 $\pm$0.097} & \textbf{0.818 $\pm$0.034}       &     \textbf{0.727 $\pm$0.013}     & \textbf{0.783 $\pm$0.047} &      \textbf{0.812 $\pm$0.051}     \\  \midrule
\begin{tabular}[c]{@{}l@{}}\textbf{FrontierX: LLM-MG } \\ --- w/o $y_{\text{exp}}$\end{tabular} & 0.664 $\pm$0.058 & 0.537 $\pm$0.071 & 0.675 $\pm$0.042       &     0.543 $\pm$0.035     & 0.673 $\pm$0.021 &      0.642 $\pm$0.075     \\  \midrule
\begin{tabular}[c]{@{}l@{}}\textbf{FrontierX: LLM-MG} \\ --- w/o $y_{\text{org}}$\end{tabular}  & 0.593 $\pm$0.027 & 0.496 $\pm$0.068 & 0.612 $\pm$0.086       &     0.482 $\pm$0.037     & 0.575 $\pm$0.076 &      0.576 $\pm$0.059     \\  \bottomrule
\bottomrule
\end{tabular}
\end{adjustbox}
\label{ablation2}
\end{table}



\bibliographystyle{plain}
\bibliography{reference}

\end{document}